\documentclass[
]{ceurart}

\usepackage{graphicx}
\usepackage{geometry}
\usepackage{caption}
\usepackage{subcaption}
\usepackage{tabularx}

\begin{document}

\copyrightyear{2024}
\copyrightclause{Copyright for this paper by its authors.
  Use permitted under Creative Commons License Attribution 4.0
  International (CC BY 4.0).}

\conference{AI for Access to Justice Workshop at ICAIL2025}

\title{In With the Old: Enhancing ‘Classical’ Document Automation with Generative AI}



\author[1]{Marc Lauritsen}[%
orcid=,
email=marc@capstonepractice.com,
]
\cormark[1]
\address[1]{Capstone Practice Systems, Harvard, Massachusetts}

\author[2]{Hannes Westermann}[%
orcid=0000-0002-4527-7316,
email=hannes.westermann@maastrichtuniversity.nl,
]
\address[2]{Maastricht Law and Tech Lab, Maastricht University, Maastricht, Netherlands}

\cortext[1]{Corresponding author.}

\begin{abstract}
   Software-based legal assistance systems have leveraged many different forms of knowledge representation and reasoning. This article explores how document automation services rooted in expert system style and other symbolic approaches can usefully enhance and be enhanced by current generative AI approaches. We discuss the possible benefits and challenges, and report on preliminary experiments in using large language models to identify and fix issues in texts written by laypeople.
\end{abstract}


\maketitle

\section{Introduction}

Legal document automation tools have been readily available since the 1980s. Hundreds of vendors have come and gone. The technology is mature and powerful, albeit seen as increasingly ‘old school’ as new forms of AI have emerged. It has natural applicability to contexts within which people without easy access to lawyers need to make decisions and produce documents with legal significance.

One service in the United States (LawHelp Interactive) has used ‘classical’ document automation (mostly HotDocs) to deliver over a million bespoke document packages each of the last several years (without charge) for people in need of legal assistance. Nonprofits and law schools have mounted many similar efforts. See \textit{Roads, Towers, and Online Legal Help} \cite{Lauritsen2020Roads} for an overview of the state of things as of several years ago.

Let’s consider how the old and new might be combined to skyrocket the effectiveness of online legal assistance. Think of it as ‘neural-symbolic.’ We explore this question both from a conceptual perspective (see sections \ref{sec:thought_experiment}, \ref{sec:benefits} and \ref{sec:challenges}) and experimentally (section \ref{sec:exp_setup} to \ref{sec:res_dis}), by carrying out an experiment on whether LLMs can detect and cure flaws in texts.

\section{Related Work}

In this paper, we discuss the integration of large language models to enhance document automation. One of the key use-cases of this approach is addressing the access to justice gap. Using AI and document assembly to address access to justice has long been a key field of study. Notable examples include the automatic assembly of protection order advisories \cite{branting2001advisory}, the providing of legal information \cite{westermann2023justicebot, westermann2023using, paquin1991loge}, dispute resolution \cite{de2021making, schmitz2021intelligent, salter2017online, bickel2015online, Branting2022-BRAACM-5} and supporting negotiation of parties. \cite{lodder2005developing, zeleznikow2016can, thompson2015creating}

In order to go beyond symbolic reasoning in such systems, we here explore the capability of large language models to analyze texts, identify flaws and produce corrected versions of narrative sections of legal forms or documents. Large language models have seen tremendous interesting in the legal field over the past few years. For example, large language models have shown promising performance in tasks such as answering legal exam questions \cite{ katz2024gpt},  statutory reasoning \cite{blair2023can}, the explanation of legal terms \cite{savelka2023explaining}, the annotation of legal documents \cite{savelka2023can, savelka2023unlocking} and the thematic analysis of legal documents \cite{drapal2023using}. 

Specifically to increase access to justice, LLMs have been explored as ways to provide legal information \cite{tan2023chatgpt, westermann2023bridging, hagan2023good}, to mediate conflicts \cite{westermann2023llmediator, tan2024robotsmiddleevaluatingllms}, to perform legal intake \cite{steenhuis2024getting} and to support the expert creation of legal information tools \cite{janatian2023text} and document automation tools \cite{steenhuis2023weaving}.

Likewise, the specific integration of large language models in document assembly has recently received attention from other researchers. \cite{Bornmann2024} explores how LLMs are able to support the drafting of documents in a domestic violence situation, with promising initial results. \cite{westermann2024dallma} introduced a system that can combine symbolic reasoning with large language models, to e.g. fill out forms or perform legal reasoning. \cite{westermann2024analyzingimageslegaldocuments} used the multi-modal capabilities of large language models to automatically extract information from textual forms, which can then be included in a document automation workflow. Here, we expand upon this work by examining whether LLMs can be given requirements in the form of text, apply these requirements to assess whether a candidate text is lacking and formulate relevant questions to obtain the missing information, and finally use the supplied information to create a corrected version of the text.

\section{A Thought Experiment}
\label{sec:thought_experiment}

\begin{figure}
    \centering
    \includegraphics[width=1\textwidth]{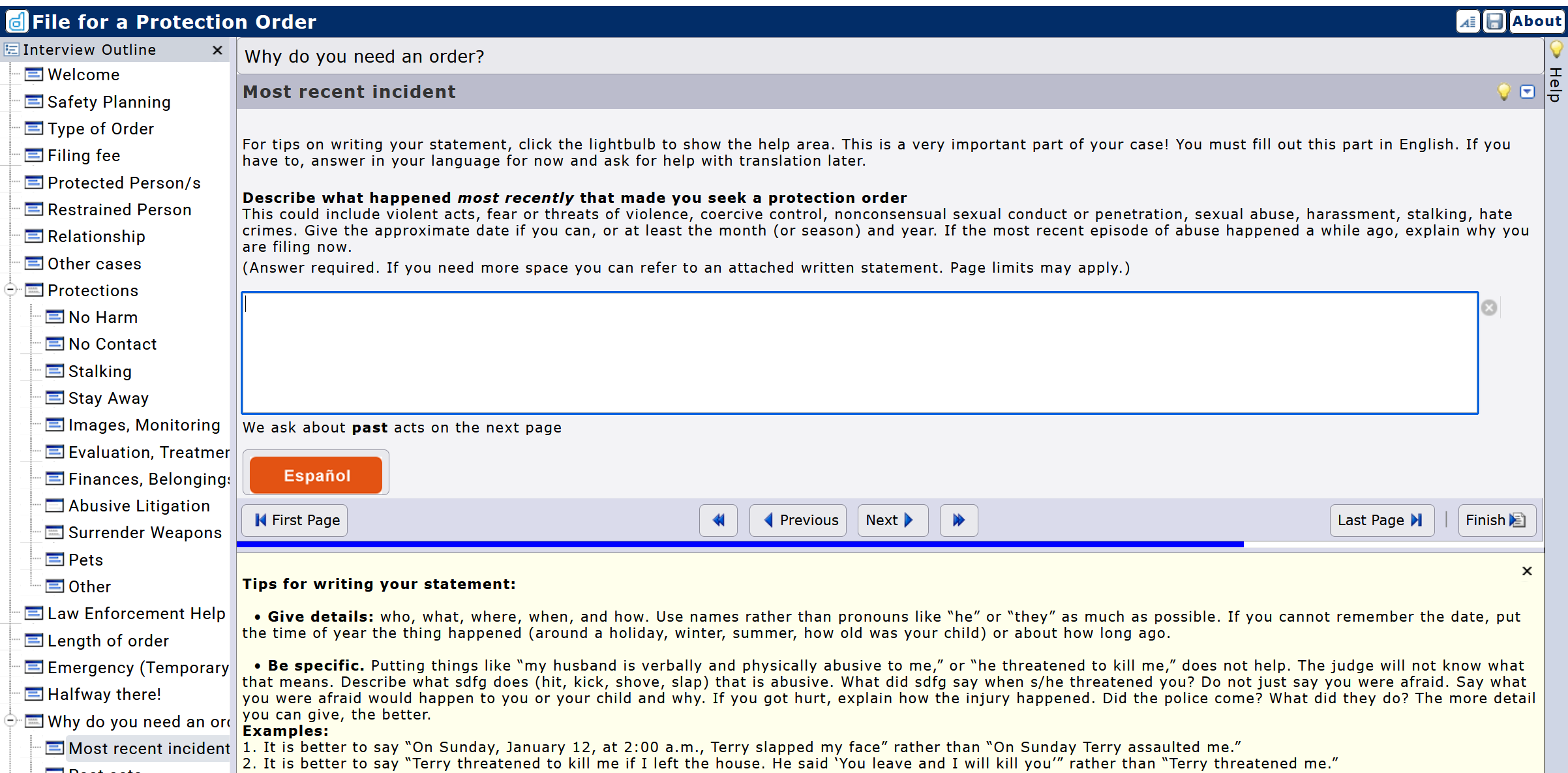}
    \caption{Screenshot of the domestic violence court order module.}
    \label{fig:module}
\end{figure}

People in Washington State who wish to get a court order against domestic violence can use this online module from Law Help Interactive (LHI).\footnote{Available online at \url{https://lawhelpinteractive.org/Interview/InterviewHome?templateId=8208}.} Upon completion, this module produces a personalized package of pleadings and instructions. Halfway through a very detailed questionnaire they see a page like the one in figure \ref{fig:module}.

Note that during the interview users can switch between English and Spanish, and there is frequent 'help' in the form of on-screen text. But users are largely left to their own devices when it comes to composing texts to enter in the empty field that calls for a long narrative response.

What if there was a button \textbf{``Click here for help in answering this question''} that brought up a conversational chat (on screen or orally) that guided the user through a process of composing an optimal factual statement?

The chat could start out by introducing itself to the user, and summarizing what they've already said in the interview. If the user hasn't already signaled her language preference by having opted to consume the interview in Spanish, it could ask ``Are you comfortable in English?'' in multiple languages, and if the answer is no, switch to whatever language they indicate.

One could draw upon answers elsewhere in the interview, since they are available in session memory on the server and/or in the local browser session and can be accessed as JSON or an XML blob.

The chat could use leading questions to gather material for a draft narrative, based on a model of what a good one looks like and interact with the user appropriately based on a model of the user that itself is based on what is already known or the user is asked. (The user might be low literacy, with little experience with the legal system, or a repeat player, perhaps a domestic violence advocate herself.) Pieces would be stitched together in a stylistically appropriate way, and then the draft text would be shown to the user - ``Does this seem accurate and complete?'' If not, or the user makes edits, the draft could be reassessed against the model. Once the process settled, the text would be entered in the interview and the user could proceed to answer other questions and assemble their documents.

A variation: LHI offers a \textit{Connect} feature for 'assisted pro se' situations in which user answers to a module like this can be reviewed and edited by a legal aid lawyer or other advocate. In that scenario, we could provide the reviewing advocate with a machine-generated assessment of the draft narrative and suggestions for improvement, which the advocate could go over with the user.

Another variation: Move the entire ‘interview’ process to a conversational one, with the LLM being driven by a definition of that interview and its requirements. (What information of what kind needs to be provided given other information that has been provided.) In other words, have the LLM ingest an interview definition (variables and logic) and use that as a script for a more flexibly interactive conversation with a user. This all could be part of more general artificial curiosity. 

Here's a quick prototype one of us cooked up, using in part some of the on-screen guidance in the above system:
\url{https://chatgpt.com/g/g-6784fd6e27208191878f923cd9dba6fa-dv-narrative-drafting}

\section{Some Benefits}
\label{sec:benefits}

Standing alone, document ‘assembly’ applications are very effective at eliciting information and producing well formatted and error-free court forms; they are not well suited for helping users compose long passages of text. Standing alone, GenAI applications can excel at question answering and conversational explorations of issues; they are not optimal for producing well formatted and error-free court forms. But together these two approaches offer compelling advantages:

\begin{itemize}
    \item By offering flexible combinations of form-like data fields and free-form exchanges, systems can address a wider spectrum of users and user needs.
    \item Rather than presenting users with long written explanations and suggestions, such content can be operationalized via system prompts to help structure interactive conversations. Such sessions can also be contextualized by information about circumstances and goals that have been gathered via more parametric affordances.
    \item Multiple languages can be made available for both inputs and outputs, as well as spoken modes.
\end{itemize}

\section{Some Challenges}
\label{sec:challenges}

Naturally, achieving these benefits will pose challenges.

\begin{itemize}
    \item First, there are the usual concerns about the quality and correctness of advice being given by an LLM-based system, especially when languages other than English are involved and users are unfamiliar with legal processes and terminologies. Workflows will need to be engineered to minimize misinformation and excessive suggestiveness.
    \item Users will need to be assured that their personal information is not being stored or used for training. Yet system providers will naturally want insight into how the systems are being used, and ideally have them ‘learn’ from such usages.
    \item Unless the classical and GenAI aspects of the user experience can be seamlessly integrated, context and interface switching can be offputting to users, and some will get lost.
    \item Someone will need to underwrite the costs of consuming sessions with the LLM provider(s), although these could perhaps be minimized by creating special purpose models that are trained on the specific requirements of the situation.
    \item Guidance-rich applications, like the one highlighted above, already come close to crossing the information/advice line in the minds of aggressive defenders of the legal profession’s monopoly on law ‘practice.’ Adding conversational aspects that help litigants compose factual declarations may be attacked as unauthorized practice, or even as ‘coaching the witness.’ See \textit{Unauthorized Help} \cite{Lauritsen2024Unauthorized} for counterarguments.
    \item Ethical issues will require attention, such as ensuring that users do not have any expectation that a lawyer-client relationship is involved and the courts and counterparties are alerted that AI assistance was made available to the litigant while they prepared their pleadings.
\end{itemize}

\section{Experimental Setup}
\label{sec:exp_setup}
In order to explore the potential of the approaches described above, we set up an experiment to validate whether an LLM can analyze a text in light of specific requirements, assess which elements are missing to comply with these requirements, request these from a user and integrate them into a complete text.

Such an approach would open the door to integrate  LLMs as drafting assistants. As part of a general document assembly system, the user could upload their initial idea of how to compose a specific narrative, upon which the LLM could help them spot issues with their text and even correct it to comply with the specific requirements of the form. This method could be very powerful, since knowing which forms to fill out and how can present a major obstacle to laypeople \cite{branting2020judges, savage2022experiences}.

Similar approaches have been attempted in areas such as protection orders \cite{Bornmann2024}, legal intake \cite{steenhuis2024getting} and in the Dallma system for legal case and letter drafting \cite{westermann2024dallma}. Here, we present an experiment that seeks to systematically evaluate the capability of an LLM to redraft a text in light of various criteria.

\subsection{Dataset}
\begin{table*}[t]
\begin{tabularx}{\textwidth}{|X|X|X|X|}
\hline
\textbf{Instructions} & \textbf{Candidate Text} & \textbf{Questions (Answers)} & \textbf{Corrected Text} \\ 
\hline
This is a draft contract for the sale of a good.   It should adhere to the following requirements or include this information: 
\begin{itemize}
    \item The name of the seller and buyer 
    \item The specific good in question 
    \item The sale price 
    \item The date of the sale 
    \item Who shall undertake the delivery 
    \item Is the good sold "as-is" or is the seller liable for a defective good? 
\end{itemize}

& I am selling a car to Reed Thomas. He will pick it up on December 20th 2024. &  
\begin{itemize}
    \item \textbf{What is the specific car being sold?}  
    1999 Volkswagen 'Bug.'
    \item \textbf{What is the sale price of the car?}  
    \$475
    \item \textbf{Is the car being sold 'as-is' or will you, as the seller, be liable for any defects?}  
    As-is
    \item \textbf{What is your name as the seller?}  
    Jane Smith
\end{itemize} &  
Contract for Sale of Goods  

This contract is entered into on December 20, 2024, by and between Jane Smith ("Seller") and Reed Thomas ("Buyer").  

1. \textbf{Description of the Good}: The Seller agrees to sell, and the Buyer agrees to purchase, a 1999 Volkswagen 'Bug.'  

2. \textbf{Sale Price}: The agreed sale price for the vehicle is \$475.  

3. \textbf{Date of Sale}: The sale shall take place on December 20, 2024.  

4. \textbf{Delivery}: The Buyer, Reed Thomas, shall undertake the responsibility of picking up the vehicle on the date of sale.  

5. \textbf{Condition of the Good}: The vehicle is sold "as-is," and the Seller, Jane Smith, makes no warranties or guarantees regarding the condition of the vehicle. The Seller shall not be liable for any defects or issues with the vehicle after the sale is completed.  

By signing below, both parties agree to the terms outlined in this contract.  
\\ 
\hline
\end{tabularx}

\caption{Example of an instructions text, a candidate text, questions (LLM) with answers provided by us, and the final text written by the LLM.}
\label{tab:dataset_overview}
\end{table*}

In order to perform the experiment, we created a dataset of three simulated scenarios, containing a textual description of requirements for a document.

\begin{enumerate}
    \item S1 - A letter reporting an issue in an apartment to a landlord. The requirements stipulate that various elements need to be included in this letter (such as the specific issue affecting the apartment and whether the landlord was contacted), as well as that the language should be formal and correct.
    \item S2 - A purchase agreement for a good. The requirements stipulate that the document needs to include elements such as the name of the buyer and seller, the good in question, the price, the date of the sale, and who is responsible for damage and delivery.
    \item S3 - A template for an employment agreement. This template is a simplified version of a real-world template for HotDocs, and as such contains text with a number of slots that need to be filled by the user (such as name of the company and employee, job title and monthly salary). The template also contains a conditional - if the employee has the right to seminar days, a specific section in the text needs to be included. Unlike the other two examples, here the expected output is much more constrained, as a specific wording is given.
\end{enumerate}

For each of these scenarios, we came up with five ``candidate texts''. These all contain one or more flaws, i.e. do not correspond to the provided document requirements. Most are missing relevant data (such as the date on which the landlord was contacted), or they may contain misspelled words. We took note of the specific flaws to facilitate evaluation.

Table \ref{tab:dataset_overview} shows the instructions for S2 and an example candidate.

\subsection{Model}
We carried out the experiments with the model \textit{gpt-4o-2024-11-20}, as a standard state-of-the-art choice. We set the temperature to 0 to reduce hallucinations and increase replicability.

\subsection{Pipeline}
In order to analyze the ability of the LLM to assess the ability of an LLM to address the issue with the candidate setup, we came up with the following two-step experimental setup:

\subsubsection{Step 1 - Generate Questions to Obtain Missing Information}

\begin{table}[h]
\centering
\begin{tabular}{|p{0.9\linewidth}|}
\hline
\textbf{INSTRUCTIONS:} \\

These are instructions on how a document should look. I will provide you with a candidate document, and you will return a list of flaws and questions to ask the user to fix these flaws. \\
\vspace{2mm}
Return a JSON object with the keys ``questions''. The value of ``questions'' is a list of strings, each representing a question that should be asked to the user to cure these flaws and capture the missing information. \\
Stick to the specific instructions. Do not ask for information that is not in the instructions. Only address the flaws with one question each. If the document is a template, only ask for the variables that are in the template. \\
\hline
\end{tabular}
\caption{Prompt for Pipeline Step 1}
\label{tab:pipeline-step1}
\end{table}

In the first step, the model is asked to generate questions to obtain the information missing from the candidate text, based on the provided instructions. The prompt (see Table \ref{tab:pipeline-step1}), thus tells the model to generate questions based on the missing information. The candidate text is then provided to the model as a user prompt.

Since the model sometimes asked too many questions, we instructed it to specifically instruct it to only ask questions actually included in the instructions. You can see an example of generated questions in table \ref{tab:dataset_overview}. Beyond this, we kept the tuning of the prompt to a minimum, to examine the behavior of a system implemented with simple instructions.

\subsubsection{Step 2 - Producing the corrected document}

Next, we wrote answers to the questions posed by the model, and examined whether it was able to use these together with the initial information to produce a corrected text, adhering to the provided instructions. The prompt for this step can be seen in Table \ref{tab:pipeline-step2}. You can see an example of a produced text in table \ref{tab:dataset_overview}.

\begin{table}[h]
\centering
\begin{tabular}{|p{0.9\linewidth}|}
\hline
\textbf{INSTRUCTIONS:} \\

These are instructions on how a document should look. I will provide you with a candidate document, as well as answers to follow-up questions asked to the user. Based on this information, create a new document that is correct, i.e. directly corresponds to the instructions. Return the answer as a text string.\\
\hline
\end{tabular}
\caption{Prompt for Pipeline Step 2}
\label{tab:pipeline-step2}
\end{table}

\subsection{Experiments}
To evaluate the ability of the LLM to ask relevant follow-up questions and integrate them into a corrected document, we examined the following questions, both quantitatively and qualitatively.

\subsection{E1 - Assessing the quality of the questions}
First, for the generated questions for a document, we assessed whether the follow-up questions asked by the model are sufficient to identify the information missing from the candidate text. We further captured notes on the process for qualitative analysis. 

Asking follow-up questions is a key feature of systems such as Dallma \cite{westermann2024dallma} and in the legal aid system \cite{steenhuis2024getting}. Here we examine whether LLMs can generate questions to reliably address flaws in documents.

\subsection{E2 - Assessing the quality of the corrected text}
After providing answers to the questions posed by the LLM, we reviewed the generated texts to investigate whether they correspond to the provided instructions. We further reviewed the output content to verify whether the answer is ``hallucination-free'', or whether the model has confabulated information. The exact questions can be seen in Table \ref{tab:quant_results}. 

Note that we took a relatively generous view on what constitutes a ``hallucination''. For a hallucination to be present, an entirely new fact would have to be introduced. Filling in details directly stemming from a fact (e.g. the loss of a laptop causing financial damages) were not counted as a hallucination, since argumentative drafting is something that LLMs could be useful for.

\section{Results \& Discussion}
\label{sec:res_dis}

\begin{table*}[h]
\centering
\begin{tabular}{|p{0.2\linewidth}|>{\centering\arraybackslash}p{0.2\linewidth}|>{\centering\arraybackslash}p{0.2\linewidth}|>{\centering\arraybackslash}p{0.2\linewidth}|}
\hline
\textbf{Scenario} & \textbf{Do the questions asked capture all of the missing elements?} & \textbf{Does the corrected text comply with the instructions?} & \textbf{Is the corrected document free of hallucinations, i.e., made up facts not part of the candidate text or questions?} \\ 
\hline
S1 - Apartment reclamation & 100\% & 100\% & 100\% \\ 
\hline
S2 - Sale Contract & 80\% & 80\% & 100\% \\ 
\hline
S3 - Employment contract & 100\% & 20\% & 100\% \\ 
\hline
\end{tabular}
\caption{Evaluation results for generated questions and corrected texts. In total, 5 examples were evaluated per scenario.}
\label{tab:quant_results}
\end{table*}

The results of our analysis are presented in table \ref{tab:quant_results}. 

\subsection{Tenant apartment reclamation}
Let us start with examining S1, regarding the tenant describing an issue in their apartment.

\textbf{Follow-up questions} - In all five cases, the model asked the questions corresponding to the flaws which we noted when creating the data. In almost all cases, the model went further than us in identifying the flaws, by asking additional questions that we had not anticipated. For example, while the instructions specified that the text should specify whether the landlord was contacted, the model also asked whether they responded and how they were contacted. On the one hand, this is good - it shows that the model takes initiative in collecting relevant information. On the other hand, it shows the difficulty of guiding LLMs, as they were specifically instructed not to go beyond the instructions in this case.

One of the requirements in the instructions was that the answer should use a formal tone, and not have any grammatical errors. The models seemed to be a bit confused as to how to deal with this requirement, and in two cases asked the user to correct their text or confirm the correct spelling of words. Perhaps, informing the model that it could later redraft the text would alleviate this issue.

\textbf{Corrected Texts} - For the corrected text, we likewise found the model to work very well. After being provided with the candidate text and the answered questions, the LLM would output a detailed text complying with the provided instructions. In all of the cases, we found that the elements lacking from the initial candidate text were included, which is a positive result. 

The style of the answer corresponded to a complete entire letter, i.e. starting with a formal greeting and ending with ``Sincerely''. This is not fully compliant with the instructions, which specify that the text represents a ``part of a submission for a claim against a landlord''. If integrating such a system into a full document assembly system, care needs to be undertaken to ensure that the generated text fits into the overall document.

Finally, while there were no direct ``hallucinations'', the model did tend to elaborate on facts that were not fully explained in the candidate text. For example, based on the information ``My computer was ruined. [due to a water leak]'', the model wrote the following text: \textit{Most notably, the leak caused damage to my computer, which is an essential item for both my personal and professional activities. This has resulted in financial loss and significant inconvenience.} 

This may be desirable in many situations, but can also be seen as an embellishment. This shows the importance of further tuning the prompt, or having a human in the loop to verify the results.

\subsection{Purchase agreement}
Next, let us examine S2, focused on the purchase agreement. Here, the requirements are more specific - while S1 includes information such as ``[the text] needs to be specific in what happened'', S2 only includes a list of elements that need to be in the text, such as the names of the sellers and buyers, the specific item sold, the price etc.

\textbf{Follow-up questions} - In S2, the model likewise did a good job of asking questions to obtain the missing information from the user, as per the instructions. In a single case, it missed that the date of sale is not necessary the same as the date of delivery. Otherwise, the questions corresponded very closely to the flaws we noted down initially.

Just like in S1, sometimes the model went further than we anticipated. For example, in one case, the context included a sentence saying that a bike would be picked up within the next two weeks. The model asked questions to attempt to clarify the specific day, and what would happen if the bike was not picked up. Even though these requirements may be legally relevant, they were not part of the instructions as provided.

\textbf{Corrected Texts} - The texts created in S2 were all found to contain all of the elements as indicated in the instructions. The model chose to adopt a specific form for the sale contract, containing an introduction specifying e.g. ``This Contract of Sale is entered into on 1/28/25 by and between ...'', and then the names of the seller and buyer. Further, the created document contains a place for the signature. These elements were not part of the instructions, but seem to stem from a general expectation of the model of what a contract of sale should look like.

None of the corrected texts contained any hallucinations, but rather stuck to the information provided.

\subsection{Employment Contract}
The final scenario, S3, focused on a template for an employment agreement. Here, the style of instructions is very different, in modeling a template document with specific slots that need to be filled. Table \ref{tab:s3_excerpt} shows an excerpt from this template. In total, the template contains 5 slots, one of which is a boolean slot that determines whether a certain section is shown or not. This conditional is provided in the template as text, to examine the ability of the LLM to apply a conditional. By being bound to a specific template, S3 is the most constrained of the scenarios.

\begin{table}[h]
\centering
\begin{tabular}{|p{0.9\linewidth}|}
\hline
EMPLOYMENT AGREEMENT\\  This Employment Agreement, by and between \{\{company\_name\}\} and \{\{employee\_name\}\} is entered into as per today.  As of today, \{\{company\_name\}\} employs \{\{employee\_name\}\}, and \{\{employee\_name\}\} accepts employment, as a full-time \{\{job\_title\}\}.\\
\hline
\end{tabular}
\caption{Part of instructions for S3}
\label{tab:s3_excerpt}
\end{table}

\textbf{Follow-up Questions} - In all five examples, the model spotted the missing information in the candidate texts, and asked the correct questions to obtain the information. However, the model was again somewhat overambitious, and sometimes asked for questions on topics that were already covered in the candidate text. 

Further, the model occasionally struggled with understanding which parts of the template were slots that could be changed (such as the name of the employee and \textit{whether} they had the right to seminar days), and also asked e.g. how many seminar days the employee had the right to and the date of the start of employment. Neither of these were slots, but rather part of the text-portion of the template. 

\textbf{Corrected text} - In general, the model was able to adhere to the instructions, and integrate the information into the template at the correct spot. The model further mostly applied the conditional correctly, by including or excluded the entire block in-between the conditional markers depending on whether we said that seminar days should be included or not.

However, the conditional did fail in one of the cases. Further, as a consequence of the additional questions for non-slots being asked, the model did frequently change details in places where they were supposed to be locked in place by the template, e.g. by adding a date or changing numbers like the employment date.

Depending on the use-case, this may not be a huge issue -- many templates serve as a drafting aid, and the user is always in control to fully change them. However, in some cases, these ``protected'' texts may be required to be intact. Here, other approaches such as e.g. semi-structured drafting \cite{westermann2024dallma}, which can guarantee the integrity of certain parts of the text, may need to be adopted.

\subsection{Overall Discussion}

Overall, the experiment shows promise in being useful for spotting missing facts, obtaining those facts from the user and using them to correct a candidate text. It is clear that this approach is powerful, and that AI systems show promise in helping e.g. laypeople craft legal narratives.

However, the experiment also shows the difficulty of ensuring precision in AI outputs. In multiple instances, the models were ``over-enthusiastic''. Instead of directly adhering to the provided instructions and prompt, the LLM would sometimes let a predefined notion of what certain types of documents should look like take precedence, especially in the absence of specific instructions regarding certain elements. While this may not always be harmful, the designers of such systems needs to be aware of this property of large language models.

This result shows that large language models are not a silver bullet for document automation. Proper care has to be used to properly specify the prompt. Further, designers of such systems may want to use approaches where the document is prepared for human review, rather than directly using the LLM output. Finally, approaches integrating other AI methods into a system (see e.g. \cite{westermann2024dallma, janatian2023text, westermann2023bridging}) may be required in certain scenarios. As always, deliberate design and evaluation of AI systems is paramount.

\bibliography{sample-ceur}

\end{document}